\documentclass[11pt]{article}

\usepackage[utf8]{inputenc}
\usepackage[T1]{fontenc}
\usepackage[margin=1in]{geometry}

\usepackage[numbers,compress]{natbib}

\usepackage{graphicx}
\usepackage{booktabs}
\usepackage{multirow}
\usepackage{array}
\usepackage{amsmath}
\usepackage{amssymb}
\usepackage{amsfonts}
\usepackage{nicefrac}
\usepackage{microtype}
\usepackage{xcolor}
\usepackage{url}
\usepackage{placeins}
\usepackage[colorlinks=true,
            linkcolor=black,
            citecolor=blue,
            urlcolor=blue]{hyperref}

\title{
\textbf{TIRA: Tumor Immune Representation Adaptation for Zero-Shot Cross-Cancer MSI and TMB Prediction}
}

\author{
\textbf{Dasari Naga Raju}
\\[0.5em]
\small Independent Researcher, Andhra Pradesh, India
\\[0.4em]
\small Corresponding Author:
\href{mailto:raajuuu1998@gmail.com}{raajuuu1998@gmail.com}
}

\date{}

\begin{document}

\maketitle

\begin{abstract}
Microsatellite instability-high (MSI-H) and high tumor mutational 
burden (TMB-H) are clinically relevant biomarkers, yet their 
histopathological prediction remains challenging when models are 
transferred across morphologically distinct cancer types. 
Immune-associated spatial patterns can persist across cancers despite 
these morphological differences, but foundation-model-based predictors 
trained on a single cancer do not explicitly use this information, 
limiting cross-cancer generalization. To address this 
limitation, we propose \textbf{TIRA} (\textbf{T}umor 
\textbf{I}mmune \textbf{R}epresentation 
\textbf{A}daptation), a target-free framework that refines 
frozen foundation-model representations using spatial immune 
topology, without requiring target-domain data during model 
development or test-time adaptation. TIRA uses a 
topology-supervised biology representation to condition tile-level 
attention while pooling only morphological features for joint MSI 
and TMB prediction. We train TIRA on TCGA-COAD+READ and evaluate it 
zero-shot on CPTAC-COAD, TCGA-STAD, TCGA-UCEC, and CPTAC-UCEC, 
covering cross-site, cross-cancer, and combined cross-cancer--site 
distribution shifts under UNI2, CONCH, and Virchow2. With UNI2, TIRA 
improved zero-shot AUROC on TCGA-STAD from 0.633 to 0.766 for MSI 
and from 0.651 to 0.772 for TMB. Source-derived spatial immune topology improved the cross-cancer 
robustness of frozen pathology foundation-model representations.
\end{abstract}
\section{Introduction}
\label{sec:intro}

Microsatellite instability-high (MSI-H) and high tumor mutational 
burden (TMB-H) are established predictive biomarkers of response to 
immune-checkpoint blockade across multiple solid tumors. Both have 
been associated with durable responses to PD-1/PD-L1 inhibition across 
different histological origins~\cite{le2017mismatch,samstein2019tmb}. Clinical assessment of MSI relies on mismatch-repair 
immunohistochemistry, PCR, or sequencing, whereas TMB requires 
sequencing-based quantification, motivating computational methods that 
jointly predict MSI and TMB from routine hematoxylin and eosin 
whole-slide images to support cost-effective triage for confirmatory 
molecular testing~\cite{kather2019deep}. Mismatch-repair deficiency can give rise to both microsatellite 
instability and elevated mutational burden, while TMB-H can also 
arise through other molecular mechanisms, reflecting the related 
but distinct biology of MSI and TMB~\cite{tcga2012crc,tcga2013ucec}. Their partially overlapping tumor--immune biology motivates joint 
prediction from a shared histological representation. The key challenge, 
however, is cross-cancer generalization: biomarker-relevant 
histological information learned from one cancer type must remain 
predictive when transferred to a morphologically distinct cancer 
without target-domain adaptation or model selection.

Large-scale pathology foundation models have enabled whole-slide 
analysis using transferable tile-level representations learned from 
diverse histological data. UNI2, CONCH, and Virchow2 employ different 
pretraining strategies and operate at different representation scales, 
providing complementary frozen feature spaces for downstream pathology 
tasks~\cite{chen2024uni,uni2repo,lu2024conch,
zimmermann2024virchow2}. These transferable feature spaces support cross-cancer prediction, 
but do not by themselves ensure that biomarker-relevant representations 
remain stable when tissue morphology changes across cancer types. Independent benchmarking has shown that the external performance of 
pathology foundation models can vary across encoders and 
evaluation cohorts~\cite{neidlinger2026benchmark}. Recent studies also 
show that their feature spaces can retain domain-specific signatures, 
which may compromise transfer when the deployment distribution differs 
from the training domain~\cite{huang2025flex}. Beyond acquisition-related variation, cross-cancer deployment 
introduces differences in organ morphology and tissue architecture, 
creating a representational mismatch that can limit zero-shot transfer. 
MSI-H tumors across different cancer types exhibit recurring 
immune-associated histological patterns. Lymphocytic infiltration, 
immune activation, and tumor-infiltrating lymphocyte accumulation 
have been reported in mismatch-repair-deficient colorectal, gastric, 
and endometrial cancers despite differences in organ-specific 
morphology~\cite{smyrk2001til,mlecnik2016immunoscore,
derosa2018gastricmsi,shia2008endometrialmsi}. Large-scale 
computational mapping of tumor-infiltrating lymphocytes from H\&E 
images across The Cancer Genome Atlas further showed that spatial 
TIL organization correlates with molecular immune features across 
cancer types~\cite{saltz2018til}. These findings motivate spatial 
immune organization as a biological prior for cross-cancer prediction. 
TMB-H reflects elevated somatic mutation burden and is also associated 
with tumor--immune interactions and response to immune-checkpoint 
blockade~\cite{samstein2019tmb}. We therefore use source-cohort spatial 
immune topology as explicit supervision for representation adaptation 
in joint MSI/TMB prediction, without assuming identical underlying 
mechanisms.

Attention-based MIL (ABMIL) uses permutation-invariant 
attention-based aggregation for bag-level prediction
~\cite{ilse2018attention}. CLAM-SB augments this framework with 
instance-level clustering constraints~\cite{lu2021data}, while 
TransMIL models correlations among instances using a transformer-based 
architecture~\cite{shao2021transmil}. ILRA exploits low-rank structure 
in slide-level MIL representations~\cite{xiang2023ilra}, and CasNet-FM 
uses cascaded MIL for histopathology-based MSI and TMB prediction
~\cite{wang2025casnet}. Although these methods improve slide-level 
aggregation, they do not explicitly address representation shift across 
cancer types. Domain-adaptation approaches can reduce distribution 
shift but often require access to target-domain samples during 
adaptation~\cite{huang2025flex}, limiting their applicability to 
zero-shot deployment. This leaves a gap for source-only methods that 
use recurring spatial immune patterns to improve the cross-cancer 
robustness of frozen foundation-model representations for joint 
MSI/TMB prediction.

The main contributions of this work are:
\begin{itemize}

\item We propose TIRA, a target-free representation adaptation 
framework for cross-cancer MSI and TMB prediction that uses 
biologically screened, source-derived spatial immune topology to 
refine frozen pathology foundation-model representations before 
deployment to unseen cancer types.

\item We develop a biology-guided MIL strategy that separates the 
roles of biology and morphology by using a topology-informed 
representation to guide tile-level attention while pooling only 
morphological features for joint biomarker prediction.
\end{itemize}

\section{Method}
\label{sec:method}

\subsection{Overview}
\begin{figure*}[t]
    \centering
\includegraphics[width=1\textwidth]{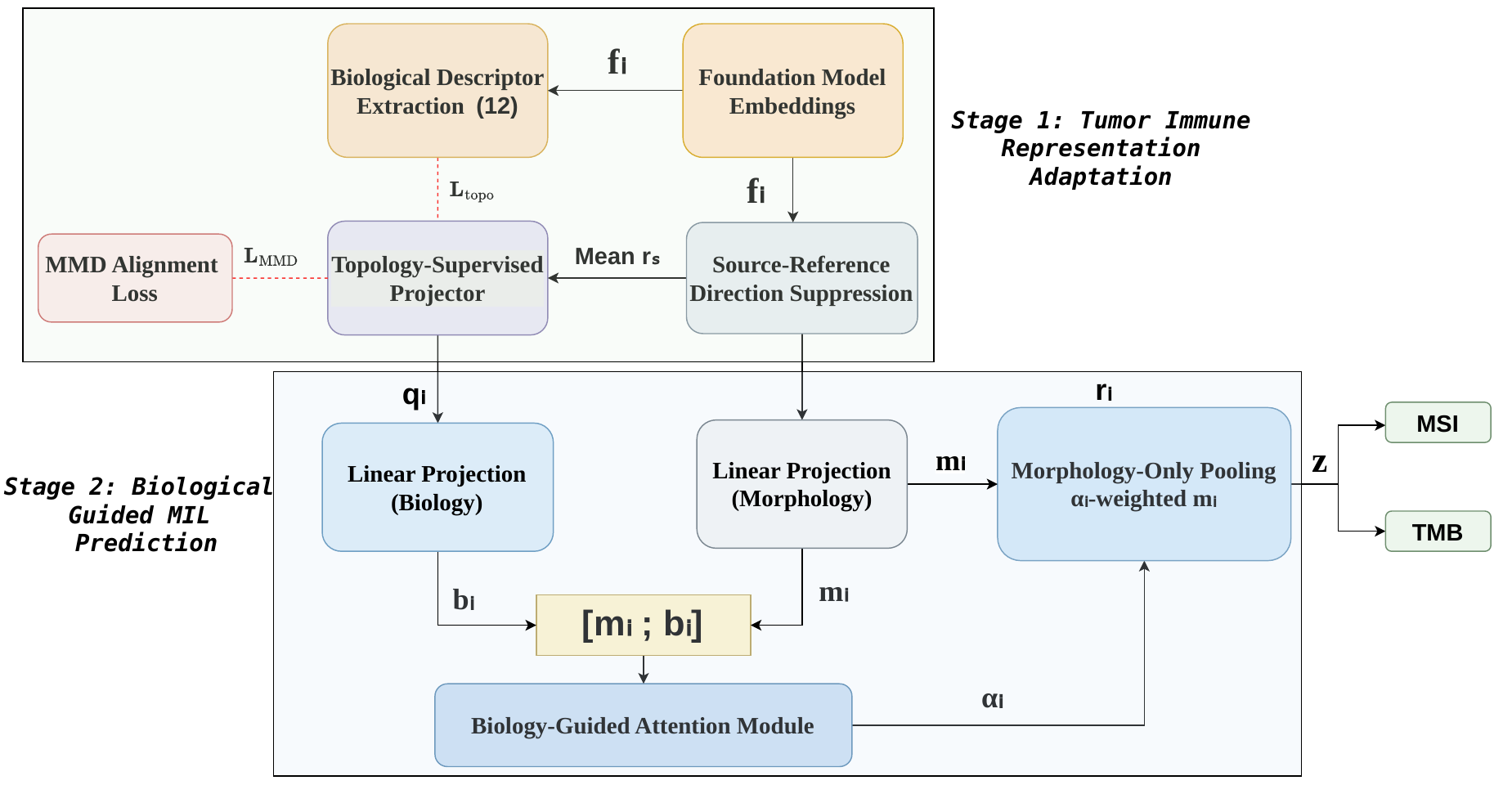}
\caption{Overview of \textbf{TIRA}. Stage~1 learns a 
topology-supervised biology representation from source data, and 
Stage~2 uses this representation to condition tile attention while 
pooling only morphological features for joint MSI/TMB prediction.}
    \label{fig:TIRA_overview}
\end{figure*}

Let a WSI be represented by $N$ tissue tiles 
$\{t_i\}_{i=1}^{N}$ with spatial coordinates 
$\mathbf{c}_i\in\mathbb{R}^2$. A frozen pathology FM $E$ 
maps each tile to $\mathbf{f}_i = E(t_i)\in\mathbb{R}^{d_E}$, 
where $d_E\in\{1536,512,2560\}$ for UNI2, CONCH, and 
Virchow2, respectively~\cite{chen2024uni,uni2repo,
lu2024conch,zimmermann2024virchow2}. TIRA consists of two stages. 
In \textbf{Stage~1}, a topology-supervised projector learns a 
biology-informed representation from slide-level mean 
direction-suppressed embeddings. In \textbf{Stage~2}, the frozen 
projector is applied tile-wise, and a biology-guided MIL predictor 
uses the resulting dual-stream representations for joint MSI/TMB 
prediction. Stage~1 learns the topology-informed representation 
without MSI/TMB supervision, while Stage~2 uses this frozen biology 
representation to guide the selection of morphological evidence for 
prediction.

\subsection{Source--Reference Direction Suppression}

We first suppress source-specific variation encoded in the FM feature 
space by estimating a unit direction from pooled tile embeddings of the 
two source subsets. For a source configuration pairing cancer cohort 
$C$ with READ as the reference,
\begin{equation}
    \mathbf{v} = 
    \frac{\boldsymbol\mu_C - \boldsymbol\mu_{\mathrm{READ}}}
    {\|\boldsymbol\mu_C - \boldsymbol\mu_{\mathrm{READ}}\|_2},
\end{equation}
where $\boldsymbol\mu_C$ and 
$\boldsymbol\mu_{\mathrm{READ}}$ are obtained by averaging all tile 
embeddings within the corresponding cohort subset. Each tile 
representation is then residualized by removing its component along 
$\mathbf{v}$:
\begin{equation}
    \mathbf{r}_i = \mathbf{f}_i - 
    (\mathbf{f}_i^\top\mathbf{v})\mathbf{v}.
    \label{eq:residual}
\end{equation}
Once estimated, $\mathbf{v}$ is frozen and applied unchanged to all 
target cohorts.

\subsection{Spatial Immune Topology}
\label{subsec:topology}

For each frozen FM, a lightweight MLP tissue classifier is trained on 
NCT-CRC-HE-100K~\cite{kather2019predicting,kather2018nct}, an auxiliary 
labeled colorectal histology dataset matching the primary source-organ 
context. From the resulting tissue-class probabilities and normalized 
tile coordinates, we compute 18 candidate spatial descriptors spanning 
tissue abundance, lymphocyte confidence and variability, local immune 
organization, peripheral and core lymphocyte distribution, multiscale 
TIL density at neighborhood scales $k\in\{10,30,100\}$, immune mixing, 
stromal infiltration, and TLS-related features. Complete definitions are 
provided in Supplementary Section~1.

\subsection{Biological Screening of Descriptors}
\label{subsec:screening}
We screen 18 candidate descriptors in 227 TCGA-COAD patients with 
matched bulk RNA-seq~\cite{tcga2012crc} using Spearman correlations 
with \textit{CD8A}, \textit{CD3E}, \textit{FOXP3}, \textit{PRF1}, 
and \textit{PDCD1}.
\begin{table}[!htb]
\centering
\caption{Source-cohort RNA-concordance screening of 18 spatial 
immune descriptors in TCGA-COAD ($n=227$). Columns report 
Spearman $\rho$ with five immune genes and their mean for UNI2.}
\label{tab:bio_validation}
\scriptsize
\setlength{\tabcolsep}{2.8pt}
\renewcommand{\arraystretch}{1.04}

\begin{tabular}{lrrrrrrc}
\toprule
& \multicolumn{5}{c}{\textbf{RNA immune-gene 
correlation}} & & \\
\cmidrule(lr){2-6}
\textbf{Descriptor} & \textbf{CD8A} & \textbf{CD3E} 
& \textbf{FOXP3} & \textbf{PRF1} & \textbf{PDCD1} 
& \textbf{Mean $\rho$} & \textbf{Use} \\
\midrule

Lymphocyte abundance    
& 0.161 & 0.188 & 0.064 & 0.013 & 0.164 
& 0.118 & $\checkmark$ \\

Tumor abundance        
& -0.168 & -0.200 & -0.180 & -0.174 & -0.110 
& -0.166 & $\times$ \\

Stromal abundance       
& 0.171 & 0.146 & 0.192 & 0.146 & 0.124 
& 0.156 & $\checkmark$ \\

Mean lymphocyte conf.   
& 0.173 & 0.201 & 0.073 & 0.027 & 0.175 
& 0.130 & $\checkmark$ \\

Lymphocyte variability  
& 0.143 & 0.177 & 0.053 & -0.005 & 0.151 
& 0.104 & $\checkmark$ \\

Local immune nbhd.      
& 0.208 & 0.253 & 0.168 & 0.106 & 0.169 
& 0.181 & $\checkmark$ \\

Local immune heterog.   
& -0.065 & -0.133 & -0.204 & -0.210 & -0.093 
& -0.141 & $\times$ \\

Peripheral lymphocytes  
& 0.173 & 0.202 & 0.072 & 0.028 & 0.184 
& 0.132 & $\checkmark$ \\

Core lymphocytes        
& 0.168 & 0.199 & 0.076 & 0.031 & 0.167 
& 0.128 & $\checkmark$ \\

Peripheral immune ratio 
& -0.018 & -0.018 & 0.004 & -0.016 & -0.001 
& -0.010 & $\times$ \\

TIL density ($k=10$)   
& 0.161 & 0.187 & 0.061 & 0.011 & 0.162 
& 0.117 & $\checkmark$ \\

TIL density ($k=30$)   
& 0.164 & 0.190 & 0.066 & 0.010 & 0.166 
& 0.119 & $\checkmark$ \\

TIL density ($k=100$)  
& 0.168 & 0.192 & 0.068 & 0.015 & 0.169 
& 0.122 & $\checkmark$ \\

Immune mixing           
& 0.183 & 0.206 & 0.074 & 0.040 & 0.182 
& 0.137 & $\checkmark$ \\

Intra-tumoral TIL       
& -0.076 & 0.030 & 0.064 & 0.004 & -0.017 
& 0.001 & $\times$ \\

Stromal TIL infiltr.   
& 0.103 & 0.067 & 0.103 & 0.168 & 0.062 
& 0.101 & $\checkmark$ \\

TLS count               
& 0.094 & 0.126 & 0.011 & -0.044 & 0.096 
& 0.057 & $\times$ \\

TLS fraction            
& 0.106 & 0.130 & 0.002 & -0.060 & 0.090 
& 0.054 & $\times$ \\

\midrule
\multicolumn{6}{r}{\textbf{Descriptors retained}} 
& \multicolumn{2}{r}{\textbf{12/18}} \\

\bottomrule
\end{tabular}
\end{table}
Descriptors with a mean correlation greater than 
0.10 across the five genes are retained, following a minimum 
positive-concordance criterion~\cite{suthen2022hypoxia}, yielding 
12 topology-supervision targets. This screening uses no MSI or TMB 
labels. The selected descriptor identities are fixed across UNI2, 
CONCH, and Virchow2, with descriptor values recomputed from each 
FM-specific tissue classifier. Table~\ref{tab:bio_validation} reports 
the UNI2 analysis, with CONCH and Virchow2 results provided in 
Supplementary Section~2.

\subsection{Stage 1: Topology-Supervised Projector}
\label{subsec:projector}

The projector $\tau_E:\mathbb{R}^{d_E}\rightarrow
\mathbb{R}^{512}$ maps direction-suppressed embeddings to a compact 
biology representation:
\begin{equation}
\tau_E(\mathbf{r}) = \operatorname{LN}\!\left(
W_2\,\operatorname{GELU}\!\left(
\operatorname{LN}(W_1\mathbf{r}+\mathbf{b}_1)
\right)+\mathbf{b}_2\right),
\label{eq:projector}
\end{equation}
where $W_1\in\mathbb{R}^{1024\times d_E}$ and 
$W_2\in\mathbb{R}^{512\times 1024}$. 
$\operatorname{LN}$ denotes layer normalization and 
$\operatorname{GELU}$ the Gaussian error linear unit
~\cite{ba2016layernorm,hendrycks2016gelu}. A topology head 
$\phi$ ($512\!\rightarrow\!256\!\rightarrow\!12$, ReLU, 
dropout~0.1) maps the projected representation to the 12 retained 
spatial descriptors.

\paragraph{Topology supervision.}
Stage~1 operates on slide-level mean residual embeddings,
$\bar{\mathbf{r}}_s=N_s^{-1}\sum_i\mathbf{r}_{si}$, with
$\bar{\mathbf{q}}_s=\tau_E(\bar{\mathbf{r}}_s)$. The retained 
descriptors are standardized across the Stage~1 source cohort before 
supervision. The topology loss is
\begin{equation}
\mathcal{L}_{\mathrm{topo}} = 
\frac{1}{|\mathcal{S}|}\sum_{s\in\mathcal{S}}
\|\phi(\bar{\mathbf{q}}_s) - \mathbf{d}_s\|_2^2,
\end{equation}
where $\mathbf{d}_s\in\mathbb{R}^{12}$ denotes the standardized 
descriptor vector for slide $s$.

\paragraph{MMD regularization.}
To reduce separation between the two source subsets in the projected 
biology space, we align their slide-level distributions using the 
biased empirical squared MMD estimator with a multi-scale RBF 
kernel~\cite{gretton2012mmd}:
\begin{equation}
\begin{aligned}
\mathcal{L}_{\mathrm{MMD}}
={}&
\frac{1}{n_C^2}\sum_{a,a'}
k(\bar{\mathbf{q}}_a,\bar{\mathbf{q}}_{a'})
+
\frac{1}{n_R^2}\sum_{b,b'}
k(\bar{\mathbf{q}}_b,\bar{\mathbf{q}}_{b'})
\\
&-
\frac{2}{n_Cn_R}\sum_{a,b}
k(\bar{\mathbf{q}}_a,\bar{\mathbf{q}}_b),
\end{aligned}
\end{equation}
where
\begin{equation}
\begin{aligned}
k(\mathbf{x},\mathbf{y})
&=
\sum_{\beta\in\mathcal{B}}
\exp\!\left(
-\frac{\|\mathbf{x}-\mathbf{y}\|_2^2}{\beta}
\right),
\\
\mathcal{B}
&=
\{0.5,1,5,10,25,50\}.
\end{aligned}
\end{equation}

The complete Stage~1 objective is
\begin{equation}
\mathcal{L}_{\mathrm{proj}} = 
\mathcal{L}_{\mathrm{topo}} + 
0.1\,\mathcal{L}_{\mathrm{MMD}}.
\label{eq:proj}
\end{equation}
The projector is optimized for 50 epochs using Adam
~\cite{kingma2015adam} (lr~$10^{-3}$, wd~$10^{-4}$), cosine 
annealing~\cite{loshchilov2017sgdr}, and batch size~32. 
Stage~1 uses no MSI or TMB labels, and $\tau_E$ is frozen after 
training.
\subsection{Stage 2: Biology-Guided Attention MIL}
\label{subsec:mil}

\paragraph{Tile-level dual-stream representation.}
Although $\tau_E$ is trained on slide-level mean embeddings, it is 
applied independently to each residual tile embedding to obtain
\begin{equation}
\mathbf{q}_i=\tau_E(\mathbf{r}_i)\in\mathbb{R}^{512}.
\end{equation}
The residual morphology feature $\mathbf{r}_i$ and topology-informed 
feature $\mathbf{q}_i$ are then independently projected to 
$\mathbf{m}_i\in\mathbb{R}^{256}$ and 
$\mathbf{b}_i\in\mathbb{R}^{256}$ using separate linear layers.

\paragraph{Biology-guided attention.}
Attention is conditioned jointly on the morphology and biology 
representations:
\begin{equation}
e_i = \mathbf{w}_a^\top\tanh\!\left(
W_a[\mathbf{m}_i;\mathbf{b}_i]+\mathbf{c}_a\right),
\quad
\alpha_i = \frac{\exp(e_i)}{\sum_j\exp(e_j)},
\label{eq:attention}
\end{equation}
where $W_a:\mathbb{R}^{512}\rightarrow\mathbb{R}^{256}$. 
The resulting attention weights are used to pool only the 
morphological representation:
\begin{equation}
\mathbf{z} = \sum_{i=1}^{N}\alpha_i\mathbf{m}_i.
\label{eq:pool}
\end{equation}
Thus, unlike standard ABMIL~\cite{ilse2018attention}, TIRA uses the 
topology-informed representation to guide tile selection while 
retaining morphology as the predictive representation.

\paragraph{Joint biomarker prediction.}
The pooled morphology vector is passed through a shared 
$256\!\rightarrow\!256$ layer (ReLU, dropout~0.25), followed by 
separate MSI and TMB classification heads. The joint objective is
\begin{equation}
\mathcal{L}_{\mathrm{MIL}} = 
\ell_{\mathrm{BCE}}(\hat{y}_s^M, y_s^M) + 
m_s^T\,\ell_{\mathrm{BCE}}(\hat{y}_s^T, y_s^T),
\label{eq:mil}
\end{equation}
where $m_s^T$ masks patients without an available TMB label. 
Both heads use unweighted binary cross-entropy.

Stage~2 is trained with Adam (lr~$10^{-4}$, wd~$10^{-4}$) and cosine 
annealing for at most 50 epochs using five MSI-stratified patient-level 
folds (seed~42) and early-stopping patience~15. For each fold, the 
checkpoint with the highest validation MSI AUROC is used for both 
biomarker heads, and source predictions are generated out-of-fold.

\section{Experimental Setup}
\label{sec:experiments}

\subsection{Cohorts and Labels}
\label{subsec:datasets}

We train on TCGA-COAD+READ~\cite{tcga2012crc} and evaluate 
zero-shot on CPTAC-COAD~\cite{vasaikar2019cptaccrc}, 
TCGA-STAD~\cite{tcga2014stad}, TCGA-UCEC~\cite{tcga2013ucec}, 
and CPTAC-UCEC~\cite{dou2020cptacucec}, covering cross-site, 
cross-cancer, and combined cross-cancer--site shifts. For reciprocal 
transfer, TCGA-STAD+READ and TCGA-UCEC+READ are used as alternative 
source configurations. The 12-descriptor topology set remains fixed 
from the initial TCGA-COAD RNA-concordance screening and is not 
reselected for these experiments.

MSI labels and TMB measurements are obtained from the corresponding 
TCGA and CPTAC molecular annotations
~\cite{tcga2012crc,tcga2014stad,tcga2013ucec,
vasaikar2019cptaccrc,dou2020cptacucec}. TMB-H is defined as 
$\geq 10$ mutations/Mb~\cite{marabelle2020tmb}.

\subsection{Foundation Models, Baselines, and Evaluation}
\label{subsec:baselines}

We evaluate frozen UNI2, CONCH, and Virchow2 representations
~\cite{chen2024uni,uni2repo,lu2024conch,zimmermann2024virchow2}. 
The MIL baselines are ABMIL~\cite{ilse2018attention}, 
CLAM-SB~\cite{lu2021data}, TransMIL~\cite{shao2021transmil}, 
CasNet-FM~\cite{wang2025casnet}, and ILRA~\cite{xiang2023ilra}. 
All methods use the same patient partitions, frozen FM features, 
and a maximum input budget of 8{,}000 tiles per bag. Stage~2 and 
all MIL baselines are trained for at most 50 epochs with learning 
rate $10^{-4}$, early-stopping patience~15, and seed~42. Each patient is represented by one precomputed tile-embedding bag. 
Source predictions are generated from the held-out fold, whereas 
external predictions average the five source-fold models. Source 
results therefore reflect cross-validated Stage~2 performance rather 
than fully nested out-of-fold evaluation of the complete two-stage 
pipeline.
\begin{table*}[t]
\centering
\caption{Patient-level MSI prediction AUROC (95\% bootstrap CI) 
for models trained on TCGA-COAD+READ and evaluated 
zero-shot on external cohorts spanning cross-site, 
cross-cancer, and combined shift.}
\label{tab:msi_results}
\footnotesize
\setlength{\tabcolsep}{3.5pt}
\resizebox{\textwidth}{!}{%
\begin{tabular}{llccccc}
\toprule
& & \textit{Source CV} & \textit{Cross-site} & \multicolumn{2}{c}{\textit{Cross-cancer}} & \textit{Cross-cancer + site} \\
\cmidrule(lr){4-4}\cmidrule(lr){5-6}\cmidrule(lr){7-7}
FM & Method 
& \begin{tabular}[c]{@{}c@{}}COAD+READ\\($n$=443, MSI-H=64)\end{tabular} 
& \begin{tabular}[c]{@{}c@{}}CPTAC-COAD\\($n$=75, MSI-H=15)\end{tabular} 
& \begin{tabular}[c]{@{}c@{}}TCGA-STAD\\($n$=308, MSI-H=54)\end{tabular} 
& \begin{tabular}[c]{@{}c@{}}TCGA-UCEC\\($n$=297, MSI-H=105)\end{tabular} 
& \begin{tabular}[c]{@{}c@{}}CPTAC-UCEC\\($n$=95, MSI-H=25)\end{tabular} \\
\midrule
\multirow{6}{*}{UNI2}
& ABMIL     & 0.8778 (0.8244--0.9244) & \textbf{0.8022 (0.6628--0.9184)} & 0.6333 (0.5458--0.7210) & 0.5153 (0.4437--0.5870) & 0.4314 (0.2868--0.5822) \\
& CLAM-SB   & 0.8715 (0.8136--0.9232) & 0.7089 (0.5675--0.8364) & 0.6490 (0.5628--0.7326) & 0.5061 (0.4358--0.5771) & 0.4829 (0.3414--0.6260) \\
& TransMIL  & 0.8861 (0.8369--0.9289) & 0.7889 (0.6487--0.9100) & 0.6395 (0.5489--0.7281) & 0.5215 (0.4486--0.5937) & 0.5069 (0.3721--0.6401) \\
& CasNet-FM & 0.8543 (0.7871--0.9151) & 0.7489 (0.6094--0.8712) & 0.6492 (0.5632--0.7340) & 0.5116 (0.4427--0.5823) & 0.4543 (0.3161--0.5948) \\
& ILRA      & 0.8361 (0.7675--0.8975) & 0.4489 (0.2804--0.6195) & 0.6457 (0.5550--0.7324) & 0.4887 (0.4187--0.5605) & 0.5023 (0.3673--0.6370) \\
& \textbf{TIRA (Ours)} & \textbf{0.8928 (0.7821--0.9372)} & 0.7932 (0.5949--0.8888) & \textbf{0.7664 (0.6934--0.8331)} & \textbf{0.5945 (0.5242--0.6352)} & \textbf{0.5606 (0.4198--0.5997)} \\
\midrule
\multirow{6}{*}{CONCH}
& ABMIL     & 0.8883 (0.8314--0.9361) & 0.6400 (0.4683--0.8036) & 0.5365 (0.4469--0.6269) & 0.5046 (0.4338--0.5746) & 0.5811 (0.4508--0.7056) \\
& CLAM-SB   & 0.8748 (0.8184--0.9224) & 0.5978 (0.4133--0.7692) & 0.5828 (0.4931--0.6716) & 0.5246 (0.4555--0.5924) & 0.6200 (0.4892--0.7430) \\
& TransMIL  & 0.8340 (0.7739--0.8890) & 0.6533 (0.5062--0.7904) & 0.5996 (0.5103--0.6861) & 0.5582 (0.4901--0.6275) & 0.5069 (0.3871--0.6242) \\
& \textbf{CasNet-FM} & \textbf{0.8909 (0.8358--0.9369)} & 0.6811 (0.5322--0.8220) & 0.5569 (0.4624--0.6491) & 0.5362 (0.4663--0.6057) & 0.6029 (0.4726--0.7260) \\
& ILRA      & 0.8620 (0.7999--0.9169) & 0.6122 (0.4597--0.7518) & 0.5337 (0.4361--0.6320) & 0.5281 (0.4572--0.5954) & 0.4709 (0.3322--0.6089) \\
& \textbf{TIRA (Ours)} & 0.8886 (0.8402--0.9298) & \textbf{0.7056 (0.5708--0.8266)} & \textbf{0.6116 (0.5257--0.6952)} & \textbf{0.5838 (0.5151--0.6525)} & \textbf{0.6291 (0.4994--0.6987)} \\
\midrule
\multirow{6}{*}{Virchow2}
& ABMIL     & 0.8598 (0.8112--0.9037) & 0.7833 (0.6376--0.9055) & 0.6911 (0.6148--0.7653) & 0.5393 (0.4683--0.6108) & 0.4851 (0.3415--0.6280) \\
& CLAM-SB   & 0.8672 (0.8134--0.9133) & 0.8100 (0.6711--0.9200) & 0.7068 (0.6266--0.7824) & 0.5599 (0.4902--0.6292) & 0.5714 (0.4394--0.7030) \\
& TransMIL  & 0.7540 (0.6931--0.8108) & 0.7867 (0.6734--0.8879) & 0.5855 (0.4968--0.6713) & 0.5823 (0.5136--0.6508) & 0.4229 (0.2941--0.5538) \\
& CasNet-FM & 0.8636 (0.8081--0.9119) & 0.7978 (0.6557--0.9127) & 0.7097 (0.6338--0.7819) & 0.5442 (0.4747--0.6124) & 0.5200 (0.3853--0.6534) \\
& ILRA      & 0.8409 (0.7813--0.8935) & 0.7333 (0.5629--0.8761) & 0.7006 (0.6262--0.7717) & 0.5229 (0.4536--0.5920) & 0.5657 (0.4363--0.6920) \\
& \textbf{TIRA (Ours)} & \textbf{0.8908 (0.8416--0.9328)} & \textbf{0.8367 (0.7035--0.9456)} & \textbf{0.7126 (0.6190--0.7829)} & \textbf{0.5901 (0.4492--0.6597)} & \textbf{0.5989 (0.4358--0.7598)} \\
\bottomrule
\end{tabular}%
}
\end{table*}
\begin{table*}[!htb]
\centering
\caption{Patient-level TMB prediction AUROC (95\% bootstrap CI) 
for models trained on TCGA-COAD+READ and evaluated 
zero-shot on external cohorts spanning cross-site, 
cross-cancer, and combined shift.}
\label{tab:tmb_results}
\footnotesize
\setlength{\tabcolsep}{3.5pt}
\resizebox{\textwidth}{!}{%
\begin{tabular}{llccccc}
\toprule
& & \textit{Source CV} & \textit{Cross-site} & \multicolumn{2}{c}{\textit{Cross-cancer}} & \textit{Cross-cancer + site} \\
\cmidrule(lr){4-4}\cmidrule(lr){5-6}\cmidrule(lr){7-7}
FM & Method 
& \begin{tabular}[c]{@{}c@{}}COAD+READ\\($n$=428, TMB-H=62)\end{tabular} 
& \begin{tabular}[c]{@{}c@{}}CPTAC-COAD\\($n$=75, TMB-H=15)\end{tabular} 
& \begin{tabular}[c]{@{}c@{}}TCGA-STAD\\($n$=308, TMB-H=56)\end{tabular} 
& \begin{tabular}[c]{@{}c@{}}TCGA-UCEC\\($n$=213, TMB-H=36)\end{tabular} 
& \begin{tabular}[c]{@{}c@{}}CPTAC-UCEC\\($n$=95, TMB-H=32)\end{tabular} \\
\midrule
\multirow{6}{*}{UNI2}
& ABMIL     & 0.9180 (0.8721--0.9552) & 0.8244 (0.6862--0.9385) & 0.6509 (0.5658--0.7352) & 0.5279 (0.4275--0.6275) & 0.4807 (0.3485--0.6097) \\
& CLAM-SB   & 0.9040 (0.8604--0.9414) & 0.7900 (0.6413--0.9176) & 0.6503 (0.5623--0.7365) & 0.5367 (0.4389--0.6347) & 0.5069 (0.3750--0.6371) \\
& TransMIL  & 0.8832 (0.8329--0.9279) & 0.8300 (0.7078--0.9289) & 0.6451 (0.5558--0.7334) & 0.5323 (0.4282--0.6347) & 0.5184 (0.3883--0.6462) \\
& CasNet-FM & 0.9149 (0.8636--0.9577) & 0.8200 (0.6902--0.9272) & 0.6698 (0.5865--0.7509) & 0.5430 (0.4436--0.6419) & 0.4742 (0.3421--0.6057) \\
& ILRA      & 0.8623 (0.8032--0.9138) & 0.5189 (0.3561--0.6799) & 0.6787 (0.5955--0.7580) & 0.4735 (0.3725--0.5741) & 0.4747 (0.3486--0.6035) \\
& \textbf{TIRA (Ours)} & \textbf{0.9218 (0.8816--0.9587)} & \textbf{0.8426 (0.7188--0.9472)} & \textbf{0.7724 (0.7035--0.8376)} & \textbf{0.5869 (0.4924--0.6788)} & \textbf{0.5315 (0.4035--0.6618)} \\
\midrule
\multirow{6}{*}{CONCH}
& ABMIL     & 0.9053 (0.8618--0.9430) & 0.7233 (0.5648--0.8616) & 0.5979 (0.5105--0.6835) & 0.4931 (0.3957--0.5906) & 0.5962 (0.4717--0.7141) \\
& CLAM-SB   & 0.8681 (0.8164--0.9156) & 0.7022 (0.5578--0.8305) & 0.7217 (0.6487--0.7907) & 0.5556 (0.4622--0.6490) & 0.6200 (0.4990--0.7365) \\
& TransMIL  & 0.8383 (0.7779--0.8922) & 0.6956 (0.5598--0.8201) & 0.6636 (0.5834--0.7406) & 0.5639 (0.4689--0.6563) & 0.5293 (0.4108--0.6494) \\
& CasNet-FM & 0.9017 (0.8605--0.9390) & 0.7322 (0.5808--0.8651) & 0.6270 (0.5380--0.7147) & 0.5077 (0.4089--0.6051) & 0.6354 (0.5141--0.7506) \\
& ILRA      & 0.8695 (0.8187--0.9152) & 0.5678 (0.3884--0.7391) & 0.6002 (0.5052--0.6944) & 0.5278 (0.4311--0.6249) & 0.5218 (0.3997--0.6466) \\
& \textbf{TIRA (Ours)} & \textbf{0.9084 (0.8693--0.9449)} & \textbf{0.7411 (0.5986--0.8618)} & \textbf{0.7295 (0.6578--0.7956)} & \textbf{0.5712 (0.4768--0.6632)} & \textbf{0.6423 (0.5219--0.7536)} \\
\midrule
\multirow{6}{*}{Virchow2}
& ABMIL     & 0.8693 (0.8221--0.9123) & 0.8011 (0.6774--0.9077) & 0.7149 (0.6414--0.7860) & 0.5692 (0.4735--0.6611) & 0.5432 (0.4172--0.6688) \\
& CLAM-SB   & 0.8743 (0.8235--0.9187) & 0.8978 (0.7939--0.9767) & 0.6995 (0.6223--0.7733) & 0.5369 (0.4371--0.6344) & 0.5481 (0.4203--0.6738) \\
& TransMIL  & 0.7650 (0.7033--0.8227) & 0.7767 (0.6600--0.8816) & 0.6091 (0.5242--0.6914) & 0.6384 (0.5438--0.7286) & 0.4221 (0.3013--0.5468) \\
& CasNet-FM & 0.8895 (0.8447--0.9289) & 0.8600 (0.7654--0.9368) & 0.7015 (0.6233--0.7757) & 0.5439 (0.4479--0.6382) & 0.5412 (0.4135--0.6668) \\
& ILRA      & 0.8538 (0.7982--0.9041) & 0.7378 (0.5648--0.8825) & 0.7343 (0.6650--0.8001) & 0.5264 (0.4294--0.6239) & 0.5967 (0.4790--0.7128) \\
& \textbf{TIRA (Ours)} & \textbf{0.9048 (0.8647--0.9419)} & \textbf{0.9051 (0.8128--0.9792)} & \textbf{0.7426 (0.6738--0.8067)} & \textbf{0.6442 (0.5508--0.7326)} & \textbf{0.6049 (0.4887--0.7168)} \\
\bottomrule
\end{tabular}%
}
\end{table*}

In the primary TCGA-COAD+READ setting, representation adaptation 
and model selection do not use target-domain information. 
No target cohort is used to estimate the suppression direction, train 
the projector or MIL model, or tune decision thresholds. The tissue 
classifier is pretrained independently on NCT-CRC-HE-100K, and the 
topology descriptor set is selected only from the TCGA-COAD 
RNA-concordance analysis. The primary metric is patient-level AUROC with nonparametric 95\% 
confidence intervals estimated from 10{,}000 patient-level bootstrap 
resamples~\cite{efron1979bootstrap}. Paired model comparisons use 
10{,}000 paired bootstrap resamples to estimate AUROC differences 
and their 95\% confidence intervals. Balanced accuracy results are 
provided in Supplementary Section~5.
\section{Results}
\label{sec:results}

\subsection{Three-Foundation-Model Benchmark}
\label{subsec:main_results}

Performance across the three foundation models varied with both the 
encoder and target cohort (Tables~\ref{tab:msi_results} 
and~\ref{tab:tmb_results}). Baseline rankings on the source cohort 
and same-cancer cross-site setting were not preserved under 
cross-cancer transfer. No single MIL baseline consistently ranked 
highest across TCGA-STAD and TCGA-UCEC, indicating that transfer 
performance depends on both the foundation-model representation and 
the target cancer. TIRA improved over matched ABMIL on TCGA-STAD and TCGA-UCEC across 
all three feature spaces, indicating that the cross-cancer effect was 
not specific to a single foundation model.

Under UNI2, the cross-cancer gains were larger on TCGA-STAD than on 
TCGA-UCEC. On TCGA-STAD, MSI AUROC increased from 0.633 with matched 
ABMIL to 0.766 with TIRA ($\Delta=+0.133$), while TMB AUROC increased 
from 0.651 to 0.772 ($\Delta=+0.121$). On TCGA-UCEC, MSI increased 
from 0.515 to 0.595 ($\Delta=+0.080$) and TMB from 0.528 to 0.587 
($\Delta=+0.059$). The magnitude of improvement also varied across foundation models, 
indicating that the effect of representation adaptation depends on 
the underlying feature space. Together, these results indicate that TIRA improves robustness under 
cancer-type shift, with the extent of improvement depending on both 
the foundation model and target cohort.

\subsection{Reciprocal Cross-Cancer Transfer}
\label{subsec:reciprocal}
Reciprocal transfer was asymmetric across source--target pairs 
(Table~\ref{tab:reciprocal_generalization}). For most pairs, MSI and 
TMB changed in the same direction relative to matched ABMIL. With 
UCEC+READ as the source, TIRA improved both biomarkers on TCGA-STAD 
but reduced both on TCGA-COAD, showing that the effect of representation 
adaptation depends on the target cancer. In contrast, transfer from 
UCEC+READ to CPTAC-UCEC improved MSI while producing only a small TMB 
change, indicating a different behavior under same-cancer cross-site 
shift. TIRA therefore improved cross-cancer transfer across several source 
configurations, while the magnitude and direction of change remained 
source--target dependent.
\begin{table*}[!htb]
\centering
\caption{Reciprocal robustness analysis under UNI2: patient-level 
AUROC for matched ABMIL and TIRA with paired 95\% CIs. 
Descriptor identities are fixed from the primary TCGA-COAD 
RNA-concordance screening across all configurations.}
\label{tab:reciprocal_generalization}
\small
\setlength{\tabcolsep}{4.5pt}
\resizebox{\textwidth}{!}{
\begin{tabular}{lllcccccc}
\toprule
& & & \multicolumn{3}{c}{\textbf{MSI}} & 
\multicolumn{3}{c}{\textbf{TMB}} \\
\cmidrule(lr){4-6}\cmidrule(lr){7-9}
\textbf{Source} & \textbf{Target} & \textbf{Shift} 
& \textbf{ABMIL} & \textbf{TIRA} 
& \textbf{$\Delta$AUROC [95\% CI]} 
& \textbf{ABMIL} & \textbf{TIRA} 
& \textbf{$\Delta$AUROC [95\% CI]} \\
\midrule

COAD+READ & TCGA-STAD  & Cross-cancer        
& 0.633 & \textbf{0.766} & +0.133 [0.076, 0.193]      
& 0.651 & \textbf{0.772} & +0.121 [0.063, 0.182] \\

COAD+READ & TCGA-UCEC  & Cross-cancer        
& 0.515 & \textbf{0.595} & +0.080 [0.029, 0.131]      
& 0.528 & \textbf{0.587} & +0.059 [$-$0.026, 0.143] \\

COAD+READ & CPTAC-UCEC & Cross-cancer + site 
& 0.431 & \textbf{0.561} & +0.130 [$-$0.032, 0.297]   
& 0.481 & \textbf{0.532} & +0.051 [$-$0.081, 0.185] \\

\midrule

STAD+READ & TCGA-COAD  & Cross-cancer        
& 0.786 & \textbf{0.859} & +0.073 [0.015, 0.137]      
& 0.805 & \textbf{0.889} & +0.084 [0.028, 0.145] \\

STAD+READ & TCGA-UCEC  & Cross-cancer        
& 0.519 & \textbf{0.678} & +0.159 [0.094, 0.226]      
& 0.550 & \textbf{0.646} & +0.096 [$-$0.014, 0.204] \\

STAD+READ & CPTAC-UCEC & Cross-cancer + site 
& 0.646 & 0.637 & $-$0.009 [$-$0.087, 0.071] 
& 0.584 & 0.584 & 0.000 [$-$0.067, 0.069] \\

\midrule

UCEC+READ & TCGA-COAD  & Cross-cancer        
& 0.712 & 0.661 & $-$0.051 [$-$0.110, 0.005] 
& 0.798 & 0.727 & $-$0.071 [$-$0.124, $-$0.021] \\

UCEC+READ & TCGA-STAD  & Cross-cancer        
& 0.598 & \textbf{0.704} & +0.106 [0.052, 0.163]      
& 0.625 & \textbf{0.726} & +0.101 [0.043, 0.164] \\

UCEC+READ & CPTAC-UCEC & Cross-site          
& 0.427 & \textbf{0.725} & +0.298 [0.158, 0.437]      
& 0.595 & \textbf{0.654} & +0.059 [$-$0.040, 0.157] \\

\bottomrule
\end{tabular}
}
\end{table*}

\subsection{Ablation Study}
\label{subsec:ablation}

\paragraph{Component ablation.}
Removing topology supervision produced the largest reduction in 
cross-cancer performance for both MSI and TMB 
(Table~\ref{tab:TIRA_ablation}), indicating that the topology-derived 
biology representation is an important component of transfer across 
cancer types. Removing biology-guided attention also reduced 
cross-cancer performance, showing that the learned biological context 
contributes through the selection of morphological evidence rather 
than through direct feature pooling.
\begin{table*}[!htb]
\centering
\caption{Component ablation of TIRA (UNI2, COAD+READ source): 
patient-level AUROC across source and zero-shot cohorts, 
with CC Drop from Full denoting the reduction in mean 
STAD/UCEC AUROC relative to full TIRA.}
\label{tab:TIRA_ablation}
\scriptsize
\setlength{\tabcolsep}{3.3pt}
\renewcommand{\arraystretch}{1.03}
\resizebox{\textwidth}{!}{%
\begin{tabular}{lcccccccccccccc}
\toprule
& \multicolumn{2}{c}{Source CV} 
& \multicolumn{2}{c}{Cross-site} 
& \multicolumn{4}{c}{Cross-cancer} 
& \multicolumn{2}{c}{Cross-cancer + site} 
& \multicolumn{2}{c}{Gen. Drop $\downarrow$} 
& \multicolumn{2}{c}{CC Drop from Full} \\
\cmidrule(lr){2-3}\cmidrule(lr){4-5}
\cmidrule(lr){6-9}\cmidrule(lr){10-11}
\cmidrule(lr){12-13}\cmidrule(lr){14-15}
Model & \multicolumn{2}{c}{COAD+READ} 
& \multicolumn{2}{c}{CPTAC-COAD} 
& \multicolumn{2}{c}{TCGA-STAD} 
& \multicolumn{2}{c}{TCGA-UCEC} 
& \multicolumn{2}{c}{CPTAC-UCEC} 
& \multicolumn{2}{c}{Source $\rightarrow$ CC} 
& \multicolumn{2}{c}{Mean STAD/UCEC} \\
\cmidrule(lr){2-3}\cmidrule(lr){4-5}
\cmidrule(lr){6-7}\cmidrule(lr){8-9}
\cmidrule(lr){10-11}\cmidrule(lr){12-13}
\cmidrule(lr){14-15}
& MSI & TMB & MSI & TMB & MSI & TMB 
& MSI & TMB & MSI & TMB & MSI & TMB 
& MSI & TMB \\
\midrule
ABMIL 
& 0.878 & 0.918 & 0.802 & 0.824 
& 0.633 & 0.651 & 0.515 & 0.528 
& 0.431 & 0.481 & 0.304 & 0.329 
& 0.106 & 0.090 \\
\midrule
TIRA Full 
& \textbf{0.893} & \textbf{0.922} 
& 0.793 & \textbf{0.843} 
& \textbf{0.766} & \textbf{0.772} 
& \textbf{0.595} & \textbf{0.587} 
& \textbf{0.561} & 0.532 
& \textbf{0.212} & \textbf{0.242} 
& -- & -- \\
$-$ Biology-Guided Attention 
& 0.876 & 0.913 & 0.782 & 0.802 
& 0.665 & 0.673 & 0.526 & 0.525 
& 0.533 & \textbf{0.554} 
& 0.281 & 0.314 & 0.085 & 0.081 \\
$-$ Direction Suppression 
& 0.891 & 0.918 & \textbf{0.823} & 0.841 
& 0.682 & 0.690 & 0.520 & 0.540 
& 0.498 & 0.523 & 0.290 & 0.303 
& 0.080 & 0.065 \\
$-$ MMD 
& 0.886 & 0.917 & 0.799 & 0.809 
& 0.670 & 0.681 & 0.522 & 0.535 
& 0.477 & 0.509 & 0.290 & 0.309 
& 0.085 & 0.072 \\
$-$ Topology Supervision 
& 0.865 & 0.896 & 0.770 & 0.801 
& 0.632 & 0.634 & 0.509 & 0.530 
& 0.471 & 0.517 & 0.295 & 0.314 
& \textbf{0.110} & \textbf{0.098} \\
\bottomrule
\end{tabular}%
}
\end{table*}
 Direction suppression showed a 
different pattern: its removal improved CPTAC-COAD MSI but reduced 
performance on TCGA-STAD and TCGA-UCEC. This contrast indicates that, 
in this setting, direction suppression is more useful for cancer-type 
shift than for same-cancer site variation, while the full model combines 
these complementary mechanisms to improve cross-cancer robustness.
\paragraph{Reference-choice ablation.}
Using READ to estimate the suppression direction yielded higher AUROC 
than both the matched random-COAD reference and no suppression across 
the STAD/UCEC endpoints (Table~\ref{tab:read_ablation}). This pattern 
indicates that the effect of direction suppression depends on the 
reference used to define the removed feature-space component, rather 
than on suppression alone. Because the comparison includes only one 
matched random-COAD reference, the analysis supports sensitivity to 
reference choice but does not establish READ as uniquely optimal.
\begin{table}[!htb]
\centering
\caption{Reference-choice ablation under UNI2. READ is compared 
with matched random-COAD and no-suppression conditions; 
$\Delta$ and 95\% CIs report READ versus random-COAD using 
10{,}000 paired bootstrap resamples.}
\label{tab:read_ablation}

\footnotesize
\setlength{\tabcolsep}{2.7pt}
\renewcommand{\arraystretch}{1.10}

\begin{tabular}{@{}llcccc@{}}
\toprule
\textbf{Dataset} &
\textbf{Task} &
\textbf{No Supp.} &
\textbf{Random} &
\textbf{READ} &
\textbf{$\Delta$ READ--Random} \\
\midrule

TCGA-STAD & MSI
& 0.682
& 0.712
& \textbf{0.766}
& +0.054 {\scriptsize [+0.010,+0.100]} \\

TCGA-STAD & TMB
& 0.690
& 0.724
& \textbf{0.772}
& +0.048 {\scriptsize [+0.000,+0.090]} \\

TCGA-UCEC & MSI
& 0.520
& 0.553
& \textbf{0.595}
& +0.042 {\scriptsize [+0.001,+0.055]} \\

TCGA-UCEC & TMB
& 0.540
& 0.521
& \textbf{0.587}
& +0.066 {\scriptsize [+0.001,+0.145]} \\

\bottomrule
\end{tabular}
\end{table}

\subsection{Attention--Immune Alignment}
\label{subsec:attention_alignment}
Tumor--lymphocyte interface enrichment was used to quantify how 
biology-guided attention changes spatial evidence selection relative to 
morphology-only ABMIL on the UNI2 cross-cancer targets. Interface tiles 
are defined as tumor-classified tiles with at least one 
lymphocyte-classified neighbor among their 30 nearest spatial neighbors. 

\begin{figure*}[!htb]
\centering
\includegraphics[width=1\textwidth]{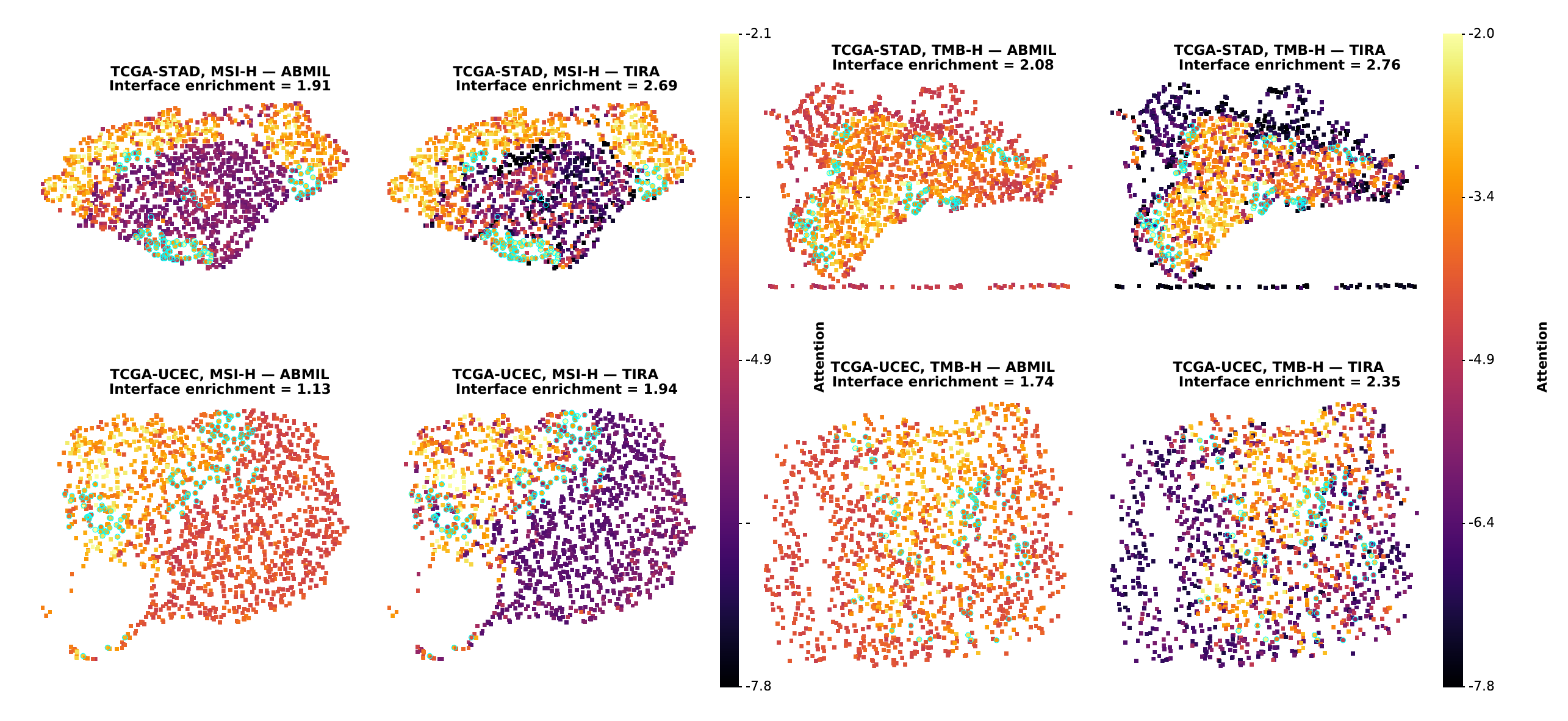}
\caption{Representative UNI2 attention maps for MSI-H and TMB-H 
cases from TCGA-STAD and TCGA-UCEC. Tiles are colored by 
$\log_{10}$ attention weight; cyan outlines denote 
tumor--lymphocyte interface tiles.}
\label{fig:attention_heatmap}
\end{figure*}

Enrichment is computed as the fraction of total attention assigned to 
interface tiles divided by their prevalence in the slide, with a value 
of 1 corresponding to uniform attention. TIRA showed higher tumor--lymphocyte interface enrichment in both 
target cohorts (Table~\ref{tab:attention_alignment}, 
Fig.~\ref{fig:attention_heatmap}). Enrichment increased from 3.34 to 
6.09 on TCGA-STAD and from 3.19 to 5.03 on TCGA-UCEC. Within the 
biomarker-specific groups, UCEC MSI-H showed the largest observed 
change, from 3.34 to 6.07, whereas the changes in STAD MSI-H and 
TMB-H were smaller. This pattern indicates that the topology-informed 
representation redistributes attention toward tumor--lymphocyte 
interfaces, with the extent of this redistribution varying across 
cohorts and biomarker subgroups. The representative maps show the 
corresponding spatial redistribution of attention.

\begin{table}[!htb]
\centering
\caption{Tumor--lymphocyte interface attention enrichment 
on UNI2 zero-shot targets; 10{,}000 paired bootstrap resamples.}
\label{tab:attention_alignment}

\footnotesize
\setlength{\tabcolsep}{3.5pt}
\renewcommand{\arraystretch}{1.05}

\begin{tabular}{llccccc}
\toprule
\textbf{Cohort} &
\textbf{Group} &
\textbf{N} &
\textbf{ABMIL} &
\textbf{TIRA} &
$\boldsymbol{\Delta}$ &
\textbf{95\% CI} \\
\midrule

TCGA-STAD & Overall 
& 308 & 3.34 & \textbf{6.09} 
& \textbf{+2.75} & \textbf{[+1.94, +3.72]} \\

TCGA-STAD & MSI-H   
& 54 & 2.03 & 2.39 
& +0.36 & {[-0.21, +0.82]} \\

TCGA-STAD & TMB-H   
& 56 & 2.02 & 2.35 
& +0.34 & {[-0.21, +0.78]} \\

\midrule

TCGA-UCEC & Overall 
& 297 & 3.19 & \textbf{5.03} 
& \textbf{+1.84} & \textbf{[+0.84, +3.54]} \\

TCGA-UCEC & MSI-H   
& 105 & 3.34 & \textbf{6.07} 
& \textbf{+2.73} & \textbf{[+0.43, +6.96]} \\

TCGA-UCEC & TMB-H   
& 36 & 4.00 & 4.74 
& +0.74 & {[-0.001, +1.46]} \\

\bottomrule
\end{tabular}

\end{table}
\subsection{Calibration Analysis}
\label{subsec:calibration}
On TCGA-STAD, TIRA reduced both ECE and Brier score relative to 
matched ABMIL for MSI and TMB (Table~\ref{tab:calibration}). The 
reduction was larger for MSI, with ECE decreasing from 0.179 to 0.063 
and Brier score from 0.184 to 0.134. On TCGA-UCEC, the calibration 
pattern differed by biomarker: TIRA improved TMB calibration, whereas 
ABMIL retained lower ECE and Brier score for MSI.
\begin{table}[!htb]
\centering
\caption{Raw probability calibration for TIRA and ABMIL on 
UNI2 zero-shot target cohorts.}
\label{tab:calibration}
\small
\setlength{\tabcolsep}{4pt}
\renewcommand{\arraystretch}{1.1}

\begin{tabular}{llcccc}
\toprule
\textbf{Cohort} & \textbf{Task} & \textbf{Model} 
& \textbf{$n$} & \textbf{ECE} & \textbf{Brier} \\
\midrule

\multirow{4}{*}{TCGA-STAD}
& \multirow{2}{*}{MSI} 
& ABMIL & 308 & 0.179 & 0.184 \\
& & TIRA & 308 & \textbf{0.063} & \textbf{0.134} \\
\cmidrule(lr){2-6}

& \multirow{2}{*}{TMB} 
& ABMIL & 308 & 0.213 & 0.199 \\
& & TIRA & 308 & \textbf{0.079} & \textbf{0.140} \\

\midrule

\multirow{4}{*}{TCGA-UCEC}
& \multirow{2}{*}{MSI} 
& ABMIL & 297 & \textbf{0.265} & \textbf{0.303} \\
& & TIRA & 297 & 0.281 & 0.318 \\
\cmidrule(lr){2-6}

& \multirow{2}{*}{TMB} 
& ABMIL & 213 & 0.171 & 0.196 \\
& & TIRA & 213 & \textbf{0.158} & \textbf{0.178} \\

\bottomrule
\end{tabular}
\end{table}

 This variation shows 
that representation adaptation can improve probability calibration in 
some transfer settings without producing the same effect across all 
cancer types and biomarkers. The cohort-level calibration pattern also parallels the discrimination 
results, where the improvement on TCGA-STAD was larger than on 
TCGA-UCEC. However, the two properties do not change uniformly under 
target-domain shift, indicating that discrimination and calibration 
should be evaluated separately. Calibration is assessed using raw 
zero-shot probabilities without post-hoc target recalibration; 
additional calibration curves and decision-curve analyses are provided 
in the Supplementary Material.

\section{Discussion}
\label{sec:discussion}

TIRA improved cross-cancer transfer across all three foundation-model 
representations, although the magnitude of improvement varied by 
encoder and target cohort. Direction suppression reduces source-associated variation, MMD aligns 
the projected source distributions, and topology supervision constrains 
the adapted representation using spatial immune descriptors screened 
for RNA concordance. Removing 
topology supervision produced the largest cross-cancer degradation, 
indicating that this biological constraint contributes to 
representation robustness under cancer-type shift. The reference-choice 
analysis further shows that the effect of direction suppression depends 
on the cohort used to estimate the suppression direction.

TIRA assigns more attention to tumor--lymphocyte interfaces in both 
TCGA-STAD and TCGA-UCEC, while only morphological features are pooled 
for MSI/TMB classification. Together with the RNA-concordance screening 
and topology ablation, this pattern supports the role of spatial immune 
topology in guiding the selection of morphological evidence rather than 
serving as a direct biomarker input. The variation in interface 
enrichment across biomarker subgroups further indicates that this 
attention guidance is context dependent and varies across cancer types 
and biomarker groups.

Reciprocal transfer showed source--target asymmetry. Using 
UCEC+READ as the source improved both biomarkers on TCGA-STAD but 
reduced performance on TCGA-COAD, showing that the effect of 
representation adaptation depends on the target cancer. This asymmetry 
may reflect differences in organ morphology, immune context, and 
biomarker distribution between source and target cohorts, although the 
current experiments do not isolate their individual contributions. TIRA 
should therefore be interpreted as improving robustness across diverse 
transfer settings rather than producing cancer-invariant 
representations. These findings support biologically informed source-only adaptation 
as a strategy for improving cross-cancer transfer without requiring 
target-domain adaptation.

\paragraph{Limitations.}
The tissue classifier used to derive spatial descriptors is trained on 
colorectal histology and applied to gastric and endometrial cohorts; 
organ-specific morphology may therefore affect tissue-class probabilities 
and downstream topology estimates. The RNA-concordance analysis provides 
patient-level biological support, but bulk RNA-seq cannot directly validate 
the spatial localization of the inferred immune patterns; spatial 
transcriptomics or immunohistochemistry would provide more direct 
validation. Reference sensitivity is evaluated using a single matched 
random-COAD pseudo-reference, and additional reference configurations are 
needed to characterize its variability more fully. The topology descriptor 
set is biologically screened in TCGA-COAD and then held fixed across 
reciprocal source configurations; broader validation is needed to determine 
whether the same descriptors remain equally informative across different 
source cancers. In the primary UNI2 comparison, cross-cancer gains on 
TCGA-UCEC were smaller than on TCGA-STAD, and the paired AUROC-difference 
confidence interval for UCEC TMB included zero. Finally, prospective 
validation is required to establish clinical utility.
\section{Conclusion}
\label{sec:conclusion}

TIRA uses source-derived spatial immune topology to adapt frozen 
foundation-model representations without target-domain adaptation. 
Across three foundation models and multiple transfer settings, TIRA 
improved the robustness of MSI and TMB prediction under cancer-type 
shift. Component ablations and attention analysis further show that 
topology-informed representations guide the selection of morphological 
evidence, particularly around tumor--immune interfaces. Spatial 
immune topology provides a source-derived biological constraint for 
improving foundation-model transfer across cancer types.

\section*{Code Availability}

The implementation of TIRA, including model training, evaluation, baseline implementations, and analysis scripts, is publicly available at
\url{https://github.com/raajuuu1998/TIRA}.

{\small
\bibliographystyle{plainnat}
\bibliography{main}
}


\clearpage

\section*{Supplementary Material}

\setcounter{section}{0}

\setcounter{figure}{0}
\setcounter{table}{0}
\setcounter{equation}{0}

\renewcommand{\thefigure}{S\arabic{figure}}
\renewcommand{\thetable}{S\arabic{table}}
\renewcommand{\theequation}{S\arabic{equation}}

\makeatletter
\renewcommand{\theHsection}{supp.\arabic{section}}
\renewcommand{\theHfigure}{supp.S\arabic{figure}}
\renewcommand{\theHtable}{supp.S\arabic{table}}
\renewcommand{\theHequation}{supp.S\arabic{equation}}
\makeatother

\section{Spatial Immune Topology Descriptor Definitions}
\label{supp:descriptor_definitions}

\paragraph{Descriptor implementation details.}
Spatial immune descriptors are computed independently for each WSI 
from FM-specific tissue-classifier probabilities and tile coordinates. 
For tile $i$, let
\begin{equation}
    \mathbf{p}_i = \operatorname{softmax}(g_E(\mathbf{f}_i))
\end{equation}
denote the tissue-class probability vector produced by the 
classifier associated with foundation model $E$, and let
\begin{equation}
    y_i = \arg\max_c p_i^{(c)}
\end{equation}
denote the corresponding hard tissue label. We use
$p_i^{\mathrm{LYM}}$, $p_i^{\mathrm{TUM}}$, and
$p_i^{\mathrm{STR}}$ for lymphocyte, tumour, and stromal
probabilities, respectively.

Tile coordinates are min--max normalised independently within each 
slide, yielding $\mathbf{c}_i\in[0,1]^2$. Let 
$\mathcal{N}_k(i)$ denote the $k$ nearest spatial neighbours of 
tile $i$. For the local immune-neighbourhood descriptors,
\begin{equation}
    k_0 = \min(10,N-1),
\end{equation}
and
\begin{equation}
\bar{p}_{i,k_0}^{(q)} =
\frac{1}{k_0}\sum_{j\in\mathcal{N}_{k_0}(i)} p_j^{(q)},
\qquad q\in\{\mathrm{LYM},\mathrm{TUM}\}.
\label{eq:supp_local_prob}
\end{equation}

The multiscale TIL-density descriptors are evaluated at
$k\in\{10,30,100\}$. Spatial tumour- and stroma-proximity
descriptors use a normalised distance threshold of $0.05$.
The stromal-TIL descriptor is evaluated when more than five
stromal tiles and at least one lymphocyte tile are available.
These constants are fixed across cohorts and experiments.

The biological screening considers 18 candidate descriptors. 
Twelve are retained for topology supervision and are denoted
$d_1,\ldots,d_{12}$ below. The remaining six candidates are
defined separately in Section~\ref{supp:excluded_candidates}.

\subsection{Lymphocyte and Stromal Abundance ($d_1$--$d_4$)}

\paragraph{Lymphocyte abundance.}
\begin{equation}
d_1 = \frac{1}{N}\sum_{i=1}^{N}\mathbf{1}[y_i=\mathrm{LYM}].
\label{eq:supp_d1}
\end{equation}

\paragraph{Stromal abundance.}
\begin{equation}
d_2 = \frac{1}{N}\sum_{i=1}^{N}\mathbf{1}[y_i=\mathrm{STR}].
\label{eq:supp_d2}
\end{equation}

\paragraph{Mean lymphocyte confidence.}
\begin{equation}
d_3 = \frac{1}{N}\sum_{i=1}^{N} p_i^{\mathrm{LYM}}.
\label{eq:supp_d3}
\end{equation}

\paragraph{Lymphocyte variability.}
\begin{equation}
d_4 = \sqrt{\frac{1}{N}\sum_{i=1}^{N}\left(p_i^{\mathrm{LYM}}-d_3\right)^2}.
\label{eq:supp_d4}
\end{equation}

\subsection{Local Immune Neighbourhood Score ($d_5$)}

\begin{equation}
q_i = \log\!\left(\frac{\bar{p}_{i,k_0}^{(\mathrm{LYM})}+\epsilon}
                       {\bar{p}_{i,k_0}^{(\mathrm{TUM})}+\epsilon}\right),
\label{eq:supp_qi}
\end{equation}
where $\epsilon$ is a small constant for numerical stability. The
slide-level descriptor is
\begin{equation}
d_5 = \frac{1}{N}\sum_{i=1}^{N}q_i.
\label{eq:supp_d5}
\end{equation}

\subsection{Peripheral and Core Lymphocyte Scores ($d_6$--$d_7$)}

Let $\bar{\mathbf{c}} = \frac{1}{N}\sum_{i=1}^{N}\mathbf{c}_i$ denote
the slide centroid. The normalised radial distance of tile $i$ is
\begin{equation}
\rho_i = \frac{\|\mathbf{c}_i-\bar{\mathbf{c}}\|_2}
              {\max_j\|\mathbf{c}_j-\bar{\mathbf{c}}\|_2}.
\label{eq:supp_radial}
\end{equation}

\paragraph{Peripheral lymphocytes.}
\begin{equation}
d_6 = \frac{1}{N}\sum_{i=1}^{N} p_i^{\mathrm{LYM}}\rho_i.
\label{eq:supp_d6}
\end{equation}

\paragraph{Core lymphocytes.}
\begin{equation}
d_7 = \frac{1}{N}\sum_{i=1}^{N} p_i^{\mathrm{LYM}}(1-\rho_i).
\label{eq:supp_d7}
\end{equation}

\subsection{Multiscale TIL Density ($d_8$--$d_{10}$)}

For each neighbourhood scale $k_\ell\in\{10,30,100\}$,
$\ell\in\{1,2,3\}$, the local TIL proportion is
\begin{equation}
u_i^{(k_\ell)} = \frac{1}{k_\ell}\sum_{j\in\mathcal{N}_{k_\ell}(i)}
\mathbf{1}[y_j=\mathrm{LYM}],
\label{eq:supp_local_til}
\end{equation}
and the slide-level descriptors are
\begin{equation}
d_{7+\ell} = \frac{1}{N}\sum_{i=1}^{N} u_i^{(k_\ell)},
\qquad k_\ell\in\{10,30,100\}.
\label{eq:supp_til_density}
\end{equation}
Thus $d_8$, $d_9$, $d_{10}$ correspond to $k=10,30,100$ respectively.

\subsection{Immune Mixing Entropy ($d_{11}$)}

\begin{equation}
u_i = \frac{1}{k_0}\sum_{j\in\mathcal{N}_{k_0}(i)}
\mathbf{1}[y_j=\mathrm{LYM}].
\label{eq:supp_mixing_fraction}
\end{equation}

Binary entropy:
$H(u)=-u\log_2u-(1-u)\log_2(1-u)$,
with $0\log_2 0\equiv0$.

\begin{equation}
d_{11} = \frac{1}{N}\sum_{i=1}^{N} H(u_i).
\label{eq:supp_d11}
\end{equation}

Higher values indicate greater local intermixing of lymphocyte- and
non-lymphocyte-labelled tissue.

\subsection{Stromal TIL Infiltration ($d_{12}$)}

Let $\mathcal{S}_{\mathrm{LYM}}=\{i:y_i=\mathrm{LYM}\}$ and
$\mathcal{S}_{\mathrm{STR}}=\{i:y_i=\mathrm{STR}\}$.
Using the fixed normalised spatial-distance threshold $\delta_s=0.05$,
\begin{equation}
d_{12} =
\frac{
\left|
\left\{
i\in\mathcal{S}_{\mathrm{LYM}}:
\min_{j\in\mathcal{S}_{\mathrm{STR}}}
\|\mathbf{c}_i-\mathbf{c}_j\|_2 < \delta_s
\right\}
\right|
}
{|\mathcal{S}_{\mathrm{LYM}}|+\epsilon}.
\label{eq:supp_d12}
\end{equation}

Evaluated when more than five stromal tiles and at least one
lymphocyte tile are available.

\subsection{Additional Candidate Descriptors Used in Screening}
\label{supp:excluded_candidates}

Six further candidates are computed during the initial biological
screening but are not used as Stage~1 topology-supervision targets.

\paragraph{Tumour abundance.}
\begin{equation}
d_{\mathrm{TUM}} =
\frac{1}{N}\sum_{i=1}^{N}\mathbf{1}[y_i=\mathrm{TUM}].
\label{eq:supp_tum_fraction}
\end{equation}

\paragraph{Local immune heterogeneity.}
\begin{equation}
d_{\mathrm{LIN\text{-}std}} =
\sqrt{\frac{1}{N}\sum_{i=1}^{N}(q_i-d_5)^2}.
\label{eq:supp_lin_std}
\end{equation}

\paragraph{Peripheral immune ratio.}
\begin{equation}
d_{\mathrm{PER\text{-}ratio}} =
\frac{d_6}{d_7+\epsilon}.
\label{eq:supp_peripheral_ratio}
\end{equation}

\paragraph{Intra-tumoral TIL density.}
Let $\mathcal{S}_{\mathrm{TUM}}=\{i:y_i=\mathrm{TUM}\}$.
\begin{equation}
d_{\mathrm{iTIL}} =
\frac{
\left|
\left\{
i\in\mathcal{S}_{\mathrm{LYM}}:
\min_{j\in\mathcal{S}_{\mathrm{TUM}}}
\|\mathbf{c}_i-\mathbf{c}_j\|_2 < 0.05
\right\}
\right|
}
{|\mathcal{S}_{\mathrm{LYM}}|+\epsilon}.
\label{eq:supp_itil}
\end{equation}

\paragraph{TLS count and TLS fraction.}
DBSCAN is applied with $\epsilon_{\mathrm{DBSCAN}}=0.05$,
$\mathrm{min\_samples}=5$.
Let $\mathcal{C}_{\mathrm{TLS}}$ denote the non-noise clusters and
$\mathcal{S}_{\mathrm{LYM}}^{\mathrm{clustered}}$ the clustered
lymphocyte tiles.

\begin{equation}
d_{\mathrm{TLS\text{-}count}} =
|\mathcal{C}_{\mathrm{TLS}}|,
\qquad
d_{\mathrm{TLS\text{-}frac}} =
\frac{|\mathcal{S}_{\mathrm{LYM}}^{\mathrm{clustered}}|}
     {|\mathcal{S}_{\mathrm{LYM}}|+\epsilon}.
\label{eq:supp_tls}
\end{equation}

\FloatBarrier

\section{Biological Validation: CONCH and Virchow2}
\label{supp:crossfm_validation}

The 12 descriptors identified under UNI2 are used as 
topology-supervision targets across all three foundation models. 
Tables~\ref{tab:conch_concordance} and~\ref{tab:virchow2_concordance} 
report RNA concordance values for CONCH and Virchow2 on the same 
18 candidates for validation purposes. Under CONCH, 12 of 18 
descriptors independently exceeded the mean Spearman $\rho > 0.10$ 
concordance criterion; under Virchow2, 11 of 18 exceeded the 
criterion, with Stromal TIL infiltration falling below threshold 
($\bar{\rho} = 0.048$). The fixed UNI2-derived descriptor set is 
retained across all foundation models to ensure a consistent 
biological supervision signal. Across both CONCH and Virchow2, 
immune-related descriptors including immune mixing, 
lymphocyte-confidence measures, and multiscale TIL density showed 
positive concordance with the selected immune genes, indicating 
that biologically relevant spatial information is retained across 
different foundation-model representations.

For each descriptor $d$, we compute its Spearman correlation with 
each of the five immune genes,
\begin{equation}
\rho_{d,g} = \operatorname{Spearman}(d,g),
\qquad
g\in\{\mathrm{CD8A},\mathrm{CD3E},\mathrm{FOXP3},
\mathrm{PRF1},\mathrm{PDCD1}\}.
\end{equation}

The mean concordance score is
\begin{equation}
\bar{\rho}_d = \frac{1}{5}\sum_{g=1}^{5}\rho_{d,g}.
\end{equation}
A descriptor passes when $\bar{\rho}_d > 0.10$.

\begin{table}[htp]
\centering
\caption{RNA-concordance for CONCH in TCGA-COAD ($n=227$).}
\label{tab:conch_concordance}
\scriptsize
\setlength{\tabcolsep}{3pt}
\renewcommand{\arraystretch}{0.95}

\begin{tabular}{lrrrrrrr}
\toprule
\textbf{Descriptor} &
\textbf{CD8A} &
\textbf{CD3E} &
\textbf{FOXP3} &
\textbf{PRF1} &
\textbf{PDCD1} &
\textbf{Mean $\rho$} &
\textbf{Pass} \\
\midrule
Lymphocyte abundance    & 0.184 & 0.234 & 0.136 & 0.113 & 0.249 & 0.183 & $\checkmark$ \\
Tumor abundance         &-0.049 &-0.065 &-0.098 &-0.024 & 0.004 &-0.047 & $\times$ \\
Stromal abundance       & 0.063 & 0.053 & 0.119 & 0.049 & 0.028 & 0.062 & $\times$ \\
Mean lymphocyte conf.   & 0.188 & 0.240 & 0.148 & 0.119 & 0.257 & 0.190 & $\checkmark$ \\
Lymphocyte variability  & 0.183 & 0.232 & 0.122 & 0.098 & 0.241 & 0.175 & $\checkmark$ \\
Local immune nbhd.      & 0.098 & 0.129 & 0.106 & 0.024 & 0.082 & 0.088 & $\times$ \\
Local immune heterog.   &-0.003 &-0.026 &-0.101 &-0.096 & 0.022 &-0.041 & $\times$ \\
Peripheral lymphocytes  & 0.160 & 0.211 & 0.108 & 0.102 & 0.231 & 0.162 & $\checkmark$ \\
Core lymphocytes        & 0.195 & 0.243 & 0.159 & 0.122 & 0.258 & 0.195 & $\checkmark$ \\
Peripheral immune ratio &-0.137 &-0.108 &-0.115 &-0.105 &-0.115 &-0.116 & $\times$ \\
TIL density ($k=10$)    & 0.183 & 0.233 & 0.130 & 0.102 & 0.244 & 0.178 & $\checkmark$ \\
TIL density ($k=30$)    & 0.186 & 0.235 & 0.134 & 0.102 & 0.248 & 0.181 & $\checkmark$ \\
TIL density ($k=100$)   & 0.195 & 0.241 & 0.139 & 0.107 & 0.255 & 0.187 & $\checkmark$ \\
Immune mixing            & 0.215 & 0.266 & 0.164 & 0.155 & 0.284 & 0.217 & $\checkmark$ \\
Intra-tumoral TIL        & 0.020 & 0.104 & 0.183 & 0.093 & 0.067 & 0.093 & $\times$ \\
Stromal TIL infiltr.     & 0.177 & 0.091 & 0.089 & 0.141 & 0.139 & 0.127 & $\checkmark$ \\
TLS count                & 0.140 & 0.158 & 0.026 & 0.027 & 0.157 & 0.102 & $\checkmark$ \\
TLS fraction             & 0.179 & 0.200 & 0.064 & 0.035 & 0.194 & 0.134 & $\checkmark$ \\
\midrule
\multicolumn{7}{r}{\textbf{Descriptors passing RNA concordance}} &
\textbf{12/18} \\
\bottomrule
\end{tabular}
\end{table}

\begin{table}[htp]
\centering
\caption{RNA-concordance for Virchow2 in TCGA-COAD ($n=227$).}
\label{tab:virchow2_concordance}
\scriptsize
\setlength{\tabcolsep}{3pt}
\renewcommand{\arraystretch}{0.95}

\begin{tabular}{lrrrrrrr}
\toprule
\textbf{Descriptor} &
\textbf{CD8A} &
\textbf{CD3E} &
\textbf{FOXP3} &
\textbf{PRF1} &
\textbf{PDCD1} &
\textbf{Mean $\rho$} &
\textbf{Pass} \\
\midrule
Lymphocyte abundance    & 0.181 & 0.185 & 0.062 & 0.065 & 0.186 & 0.136 & $\checkmark$ \\
Tumor abundance         &-0.101 &-0.132 &-0.168 &-0.055 &-0.038 &-0.099 & $\times$ \\
Stromal abundance       & 0.015 &-0.010 & 0.050 &-0.008 & 0.027 & 0.015 & $\times$ \\
Mean lymphocyte conf.   & 0.205 & 0.202 & 0.074 & 0.076 & 0.205 & 0.153 & $\checkmark$ \\
Lymphocyte variability  & 0.160 & 0.170 & 0.040 & 0.036 & 0.168 & 0.115 & $\checkmark$ \\
Local immune nbhd.      & 0.181 & 0.208 & 0.157 & 0.068 & 0.123 & 0.147 & $\checkmark$ \\
Local immune heterog.   &-0.095 &-0.120 &-0.217 &-0.205 &-0.073 &-0.142 & $\times$ \\
Peripheral lymphocytes  & 0.209 & 0.211 & 0.068 & 0.093 & 0.216 & 0.160 & $\checkmark$ \\
Core lymphocytes        & 0.183 & 0.179 & 0.072 & 0.052 & 0.177 & 0.133 & $\checkmark$ \\
Peripheral immune ratio &-0.071 & 0.028 & 0.048 & 0.022 &-0.025 & 0.030 & $\times$ \\
TIL density ($k=10$)    & 0.175 & 0.179 & 0.056 & 0.056 & 0.180 & 0.129 & $\checkmark$ \\
TIL density ($k=30$)    & 0.181 & 0.182 & 0.054 & 0.055 & 0.182 & 0.131 & $\checkmark$ \\
TIL density ($k=100$)   & 0.182 & 0.182 & 0.053 & 0.054 & 0.184 & 0.131 & $\checkmark$ \\
Immune mixing            & 0.209 & 0.209 & 0.085 & 0.095 & 0.212 & 0.162 & $\checkmark$ \\
Intra-tumoral TIL        & 0.049 & 0.097 & 0.162 & 0.054 & 0.067 & 0.086 & $\times$ \\
Stromal TIL infiltr.     & 0.028 & 0.050 & 0.086 & 0.035 & 0.040 & 0.048 & $\times$ \\
TLS count                & 0.133 & 0.144 & 0.047 & 0.050 & 0.143 & 0.103 & $\checkmark$ \\
TLS fraction             & 0.105 & 0.114 & 0.066 &-0.006 & 0.094 & 0.075 & $\times$ \\
\midrule
\multicolumn{7}{r}{\textbf{Descriptors passing RNA concordance}} &
\textbf{11/18} \\
\bottomrule
\end{tabular}
\end{table}

\FloatBarrier

\section{Zero-Shot Probability Calibration}
\label{sec:supp_calibration}

We examine raw UNI2 probability calibration on the primary 
cross-cancer targets, TCGA-STAD and TCGA-UCEC, using the same 
patient-level predictions as the zero-shot evaluation. Calibration 
curves are constructed using five quantile-based bins without 
post-hoc target-cohort recalibration. For each bin, the horizontal 
axis represents the mean predicted probability and the vertical axis 
represents the observed event frequency. Predictions closer to the 
diagonal are better calibrated, whereas points below the diagonal 
indicate overestimation of risk and points above the diagonal indicate 
underestimation.
\begin{figure}[!htp]
    \centering
    \includegraphics[width=\textwidth]{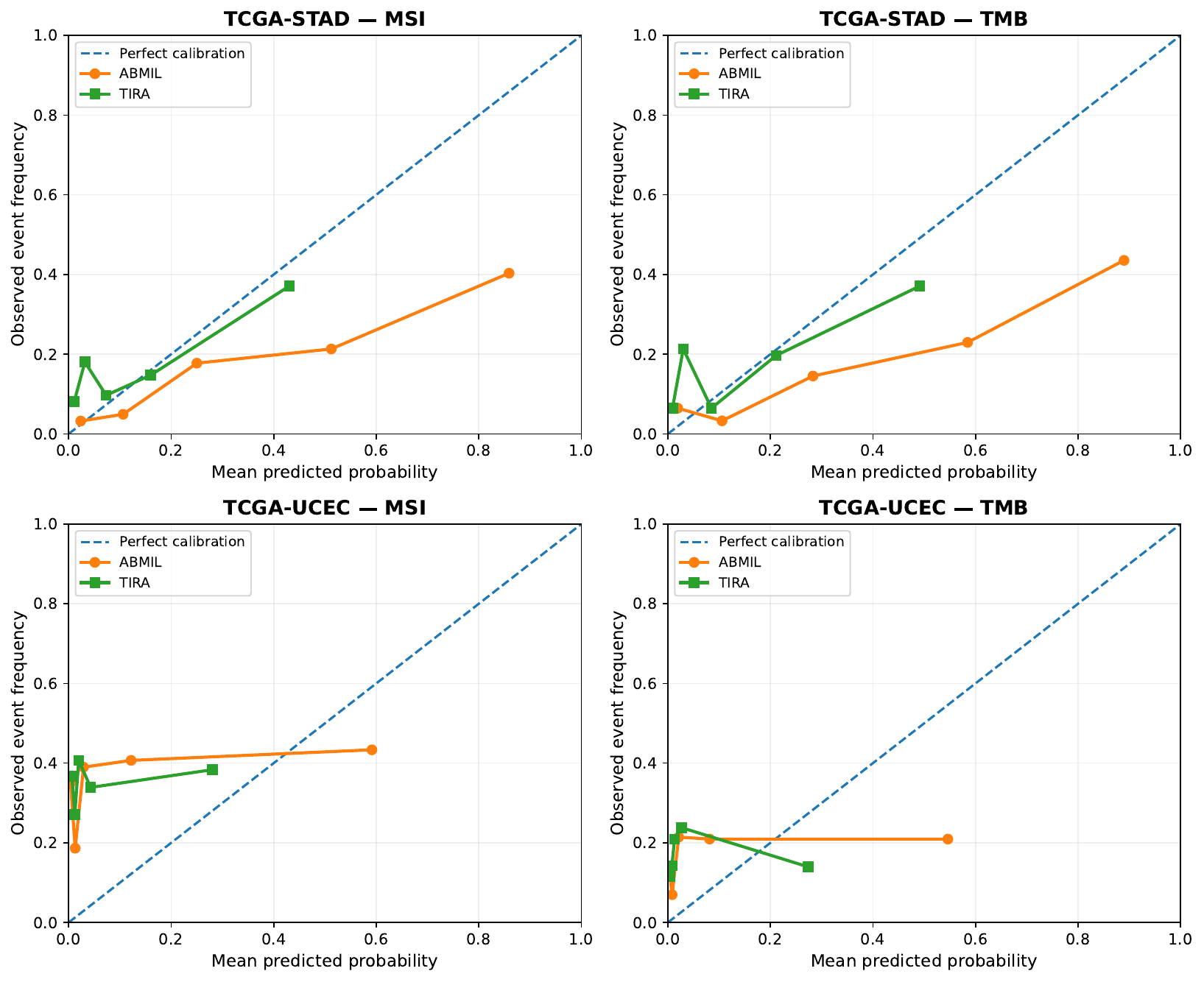}
    \caption{Zero-shot calibration of ABMIL and TIRA under UNI2 on 
    TCGA-STAD and TCGA-UCEC. Curves use five quantile-based bins; 
    the dashed diagonal denotes perfect calibration.}
    \label{fig:supp_calibration}
\end{figure}

The calibration behavior differed across target cohorts and 
biomarkers. On TCGA-STAD, the TIRA curves for both MSI and TMB 
generally remained closer to the perfect-calibration diagonal than 
those of matched ABMIL, consistent with the lower ECE and Brier 
scores reported in the main paper. In particular, ABMIL tended to 
assign relatively high probabilities to bins with substantially lower 
observed event frequencies, whereas TIRA reduced this discrepancy. 
On TCGA-UCEC, the pattern was less consistent. For MSI, ABMIL showed 
better overall calibration than TIRA, while for TMB, TIRA provided a 
modest improvement over ABMIL. Both UCEC panels nevertheless show 
substantial departures from the diagonal, indicating that probability 
estimates remain sensitive to cancer-type shift even when 
discrimination improves.

\FloatBarrier

\section{Decision-Curve Analysis}
\label{sec:supp_dca}

We further evaluate the clinical decision behavior of raw UNI2 
predictions on TCGA-STAD and TCGA-UCEC using decision-curve 
analysis. TIRA and matched ABMIL are compared with treat-all 
and treat-none strategies across threshold probabilities from 
0.01 to 0.50. Net benefit at threshold $p_t$ is defined as

\begin{equation}
\mathrm{NB}(p_t) =
\frac{\mathrm{TP}}{N}
-
\frac{\mathrm{FP}}{N}
\frac{p_t}{1-p_t}.
\end{equation}

\begin{figure}[htp]
    \centering
    \includegraphics[width=\textwidth]{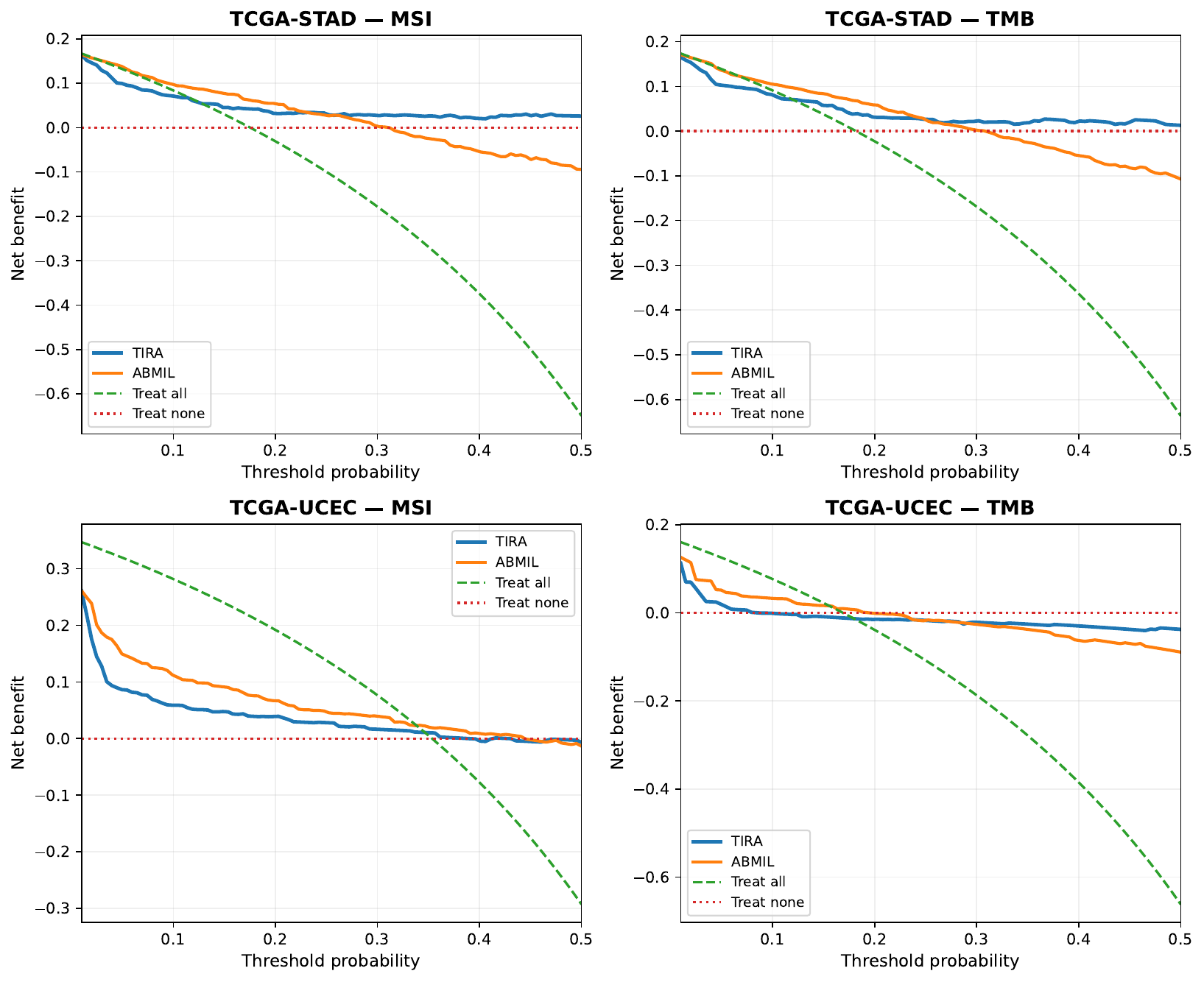}
    \caption{Decision-curve analysis of ABMIL and TIRA under UNI2 
    on TCGA-STAD and TCGA-UCEC for MSI and TMB. Net benefit is 
    shown across threshold probabilities from 0.01 to 0.50, with 
    treat-all and treat-none strategies included as references.}
    \label{fig:supp_dca}
\end{figure}

On TCGA-STAD, TIRA maintained positive net benefit across 
a broader range of moderate-to-high thresholds than ABMIL for 
both MSI and TMB. On TCGA-UCEC, net benefit was more limited 
and threshold dependent, with no consistent advantage across 
the evaluated range. These results indicate that decision-curve 
behavior under zero-shot transfer varies with the target cancer 
and decision threshold.

\FloatBarrier

\section{Additional Classification Metric}
\label{sec:supp_additional_metric}

We additionally report balanced accuracy for MSI and TMB across all three foundation models. Source-derived decision thresholds are applied unchanged to the external cohorts.

\begin{table}[htp]
\centering
\caption{Balanced accuracy for MSI prediction across foundation models and evaluation cohorts.}
\label{tab:supp_msi_balanced}
\scriptsize
\setlength{\tabcolsep}{4pt}
\renewcommand{\arraystretch}{1.03}

\begin{tabular}{llccccc}
\toprule
\textbf{FM} & \textbf{Model} &
\textbf{COAD+READ} & \textbf{CPTAC-COAD} &
\textbf{TCGA-STAD} & \textbf{TCGA-UCEC} &
\textbf{CPTAC-UCEC} \\
\midrule

\multirow{6}{*}{UNI2}
& ABMIL                & 0.823 & 0.500          & 0.566          & 0.496          & 0.477 \\
& CLAM-SB              & 0.833 & 0.492          & 0.582          & 0.513          & 0.511 \\
& TransMIL             & 0.814 & 0.608          & 0.566          & 0.514          & 0.543 \\
& CasNet-FM            & 0.844 & 0.617          & 0.544          & 0.510          & 0.489 \\
& ILRA                 & 0.798 & 0.467          & 0.598          & 0.512          & 0.501 \\
& \textbf{TIRA (Ours)} & 0.841 & \textbf{0.742} & \textbf{0.647} & \textbf{0.531} & 0.540 \\

\midrule

\multirow{6}{*}{CONCH}
& ABMIL                & \textbf{0.854} & 0.550          & 0.500          & 0.516          & 0.577 \\
& CLAM-SB              & 0.830          & \textbf{0.558} & 0.566          & \textbf{0.522} & 0.589 \\
& TransMIL             & 0.779          & 0.500          & 0.552          & 0.518          & 0.459 \\
& CasNet-FM            & 0.839          & 0.525          & 0.559          & 0.514          & 0.557 \\
& ILRA                 & 0.805          & 0.525          & 0.555          & 0.490          & 0.490 \\
& \textbf{TIRA (Ours)} & 0.848          & 0.552          & \textbf{0.582} & 0.520          & \textbf{0.593} \\

\midrule

\multirow{6}{*}{Virchow2}
& ABMIL                & 0.782          & 0.708          & 0.597          & 0.520          & 0.454 \\
& CLAM-SB              & 0.800          & \textbf{0.733} & 0.619          & 0.539          & \textbf{0.583} \\
& TransMIL             & 0.705          & 0.617          & 0.545          & \textbf{0.569} & 0.493 \\
& CasNet-FM            & 0.796          & 0.692          & 0.608          & 0.537          & 0.521 \\
& ILRA                 & 0.772          & 0.667          & 0.610          & 0.513          & 0.573 \\
& \textbf{TIRA (Ours)} & \textbf{0.810} & 0.724          & \textbf{0.623} & 0.561          & 0.579 \\

\bottomrule
\end{tabular}
\end{table}

\begin{table}[htp]
\centering
\caption{Balanced accuracy for TMB prediction across foundation models and evaluation cohorts.}
\label{tab:supp_tmb_balanced}
\scriptsize
\setlength{\tabcolsep}{4pt}
\renewcommand{\arraystretch}{1.03}

\begin{tabular}{llccccc}
\toprule
\textbf{FM} & \textbf{Model} &
\textbf{COAD+READ} & \textbf{CPTAC-COAD} &
\textbf{TCGA-STAD} & \textbf{TCGA-UCEC} &
\textbf{CPTAC-UCEC} \\
\midrule

\multirow{6}{*}{UNI2}
& ABMIL                & 0.866          & 0.625          & 0.557          & 0.473          & \textbf{0.501} \\
& CLAM-SB              & 0.833          & \textbf{0.742} & 0.571          & 0.531          & 0.485 \\
& TransMIL             & 0.815          & 0.708          & 0.565          & \textbf{0.543} & 0.494 \\
& CasNet-FM            & \textbf{0.881} & 0.558          & 0.574          & 0.468          & 0.478 \\
& ILRA                 & 0.813          & 0.550          & 0.635          & 0.457          & 0.492 \\
& \textbf{TIRA (Ours)} & 0.876          & 0.734          & \textbf{0.640} & 0.538          & 0.499 \\

\midrule

\multirow{6}{*}{CONCH}
& ABMIL                & \textbf{0.851} & 0.667          & 0.546          & 0.490          & 0.605 \\
& CLAM-SB              & 0.812          & 0.542          & 0.668          & 0.457          & \textbf{0.612} \\
& TransMIL             & 0.784          & 0.517          & 0.601          & 0.494          & 0.520 \\
& CasNet-FM            & 0.832          & 0.600          & 0.613          & 0.488          & 0.590 \\
& ILRA                 & 0.799          & 0.450          & 0.553          & \textbf{0.512} & 0.543 \\
& \textbf{TIRA (Ours)} & 0.842          & \textbf{0.671} & \textbf{0.674} & 0.506          & 0.608 \\

\midrule

\multirow{6}{*}{Virchow2}
& ABMIL                & 0.791          & 0.767          & 0.592          & 0.541          & 0.516 \\
& CLAM-SB              & 0.819          & 0.783          & 0.582          & 0.503          & 0.500 \\
& TransMIL             & 0.721          & 0.600          & 0.577          & \textbf{0.638} & 0.494 \\
& CasNet-FM            & \textbf{0.832} & 0.775          & 0.563          & 0.567          & 0.493 \\
& ILRA                 & 0.812          & 0.658          & \textbf{0.644} & 0.511          & \textbf{0.612} \\
& \textbf{TIRA (Ours)} & 0.827          & \textbf{0.787} & 0.637          & \textbf{0.642} & 0.603 \\

\bottomrule
\end{tabular}
\end{table}

\FloatBarrier

\end{document}